\documentclass[letterpaper]{article} 
\usepackage{aaai2026}
\copyrighttext{*Corresponding Author. Email: siamdip13@gmail.com}

\usepackage{times}  
\usepackage{helvet}  
\usepackage{courier}  
\usepackage[hyphens]{url}  
\usepackage{graphicx} 
\usepackage{natbib}  
\usepackage{caption} 
\usepackage{algorithm}
\usepackage{algorithmic}

\usepackage{newfloat}
\usepackage{listings}
\DeclareCaptionStyle{ruled}{labelfont=normalfont,labelsep=colon,strut=off} 
\floatstyle{ruled}
\newfloat{listing}{tb}{lst}{}
\floatname{listing}{Listing}
\usepackage{circledtext}

\usepackage[table]{xcolor}

\definecolor{summaryblue}{RGB}{235,244,250}

\title{SPeaR: Test-Time Adaptation with Steering Primitives for Realigning Representations}

\author{
    Muhammad Sudipto Siam Dip*,
    Ali Etemad
}

\affiliations{
    Department of Electrical and Computer Engineering\\
    Queen's University, Canada
}

\usepackage{bibentry}

\usepackage{booktabs} 
\usepackage{amsmath}
\usepackage{url}
\usepackage[inline]{enumitem}
\usepackage{multirow}
\usepackage{enumitem}
\usepackage{subcaption}
\usepackage{soul}
\usepackage{amssymb}

\begin{document}

\maketitle

\begin{abstract}

Test-time adaptation (TTA) addresses distribution shift using only unlabeled test data. Existing methods typically adapt pretrained models by updating their parameters, limiting both what is adapted and where adaptation can occur within the network. We instead keep the pretrained network frozen and steer its intermediate representations. We introduce SPeaR (Steering Primitive for Realigning Representations), which inserts lightweight learnable modules at stage boundaries and optimizes them directly from the test stream, requiring neither source data nor supervised warm-up. Each primitive is optimized using a gated objective that reduces uncertainty only when adaptation is beneficial, along with a diversity regularizer to prevent collapse, and a multi-depth anchor to stabilize adaptation. We show that steering early representations is the most effective strategy, and that the same primitive transfers across convolutional and Transformer architectures. Across CIFAR-10-C, CIFAR-100-C, and ImageNet-C, SPeaR consistently matches or outperforms methods that adapt orders of magnitude more parameters, remains robust across a wide range of batch sizes, and preserves source-domain performance during continual adaptation.

\end{abstract}

\section{Introduction}

Test-time adaptation (TTA) enables pretrained models to adapt to distribution shifts encountered during deployment without access to source data or target labels~\cite{wang2024search,liang2024comprehensive}. This capability is important in practice since models frequently face data distributions during deployment, which differ from those used during training. Such distribution shifts often violate the i.i.d. assumption and cause significant performance degradation
~\cite{hendrycks2019benchmarking,koh2021wilds}. Unlike domain adaptation, which requires target data during training, or domain generalization, which cannot anticipate all deployment conditions, TTA adapts the model on the fly using only the incoming test stream 
~\cite{ganin2016domain,wilson2020survey,zhou2022domain}.

Distribution shifts affect the representations throughout the network, resulting in feature statistics that deviate from those of the source domain. This deviation accumulates with depth until the representation that reaches the classifier is no longer the one the decision boundary was originally trained for. Most existing TTA methods, however, address this by updating model weights, whether by recalibrating normalization layers \cite{wang2020tent, niu2022efficient, lee2024continual}, fine-tuning the full network \cite{wang2022continual, zhang2022memo}, or selecting a subset of layers to adapt \cite{sahoo2025layer, ambekar2026hierarchical}. Updating weights to correct a representation is indirect: it alters the pretrained solution itself, which makes adaptation vulnerable to error accumulation and catastrophic forgetting, and ties the method to wherever trainable parameters are available in the architecture. A smaller body of work instead corrects the representation with inserted modules while keeping the backbone frozen (Song et al. 2023; Shin and Kim 2024), but these modules must be warm-started with supervised training on source data, which conflicts with the source-free assumption of TTA, or they introduce parallel branches with orders of magnitude more parameters (Kim, Han, and Hwang 2025). It thus remains open whether a shifted representation can be realigned directly, using only the unlabeled test stream, without modifying any pretrained weight or preparing anything in advance. Answering this raises two questions that weight-based adaptation does not encounter: at which depth should such a correction be applied, and what should it aim to restore?

In this paper, we introduce SPeaR (\textbf{S}teering \textbf{P}rimitive for R\textbf{ea}ligning \textbf{R}epresentations) to address both questions. SPeaR is a lightweight test-time adaptation framework that steers intermediate representations while keeping the pretrained backbone completely frozen. SPeaR inserts simple representation steering primitives, which we name Representation Steering Primitives (RSP), at selected network stages and optimizes them directly from the unlabeled test stream, requiring neither source data nor supervised warm-up. The RSP parameters are optimized using a gated objective that reduces prediction uncertainty only when adaptation is necessary, a diversity term to prevent class collapse, and an anchor objective that discourages unnecessary deviations from the identity transformation. This allows SPeaR to introduce only the representation corrections supported by the incoming test data while maintaining stable adaptation throughout deployment.
Through comprehensive experiments, we show that effective adaptation is better achieved by steering representations early on in the network, where small corrections propagate through subsequent layers to recover robust downstream features. Building on this observation, we introduce a hierarchical anchoring strategy that stabilizes multi-depth adaptation by accounting for the different optimization dynamics of shallow and deep steering primitives. Across diverse architectures, including batch-normalized, group-normalized, and normalization-free networks, and multiple test-time adaptation settings, SPeaR consistently achieves state-of-the-art or competitive performance while optimizing only a very small number of parameters and leaving every pretrained weight untouched.

In summary, our contributions are as follows:
\textbf{1.}~
\textbf{Representation steering without source-side warm-up.}  We introduce SPeaR, which realigns intermediate representations using a steering primitive inserted into a frozen network. The primitive begins at the identity and is recovered entirely from the unlabeled test stream, requiring no source data, no labels, no auxiliary controller, and no supervised warm-up.
\textbf{2.}~\textbf{SPeaR is extremely efficient.} SPeaR operates by adapting as few as 64 parameters (when inserted in a network with $\sim$1.4 million parameters) and up to only $\sim$4.6K parameters (when inserted in a network with $\sim$86 million parameters). This is orders of magnitude lower than state-of-the-art solutions. At the same time, our method achieves better or competitive performance compared to prior works.
\textbf{3.}~\textbf{SPeaR is architecture-agnostic.} Unlike most prior methods that rely on normalization layers or architecture-specific adaptation mechanisms, SPeaR operates through the same steering primitive across batch-normalized, group-normalized, normalization-free, and even Transformer networks, enabling a single adaptation framework for diverse backbone architectures.
\textbf{4.}~\textbf{Continual adaptation and robustness to forgetting.} SPeaR substantially mitigates catastrophic forgetting while maintaining robust performance under continual distribution shifts by keeping the backbone frozen, optimizing only lightweight steering primitives, and regularizing the RSP optimization with an anchor objective.
\textbf{5.}~\textbf{Reproducibility and code release.} We have provided the complete implementation to facilitate reproducible research and future development.

\section{Related Works}
\label{sec:relatedworks}

\noindent \textbf{TTA w/ Model Parameters Updates.} 
Adaptation via updating model parameters is performed by either updating normalization layers \cite{wang2020tent, niu2022efficient, niu2023towards, marsden2024universal, lee2024continual, lee2024entropy}, updating the full network \cite{wang2022continual, zhang2022memo} or restricting the update to selected layers \cite{sahoo2025layer, ambekar2026hierarchical}. Alternatively, some methods leave parameters fixed and interpolate source and test-batch statistics instead \cite{schneider2020improving, khurana2021sita}, with TTN \cite{lim2023ttn} calibrating the coefficient offline on labeled source data.

\noindent \textbf{TTA w/ Auxiliary Component Adaptation.} A growing line of work shifts focus from how to adapt to what to adapt, introducing dedicated trainable modules while keeping the backbone frozen. RNA \cite{yeo2023rapid} inserts adapters into a frozen network, but does not optimize them at test time; instead it trains a controller on labeled source data to predict them in a single forward pass from an external adaptation signal. EcoTTA \cite{song2023ecotta} and L-TTA \cite{shin2024tta} follow this direction, but require source data to initialize their modules, unlike the source-free setting that TTA assumes. Buffer Layers \cite{kim2025buffer} remove this requirement and adapt inserted modules without source access. However, they require intrusive modifications to the model's forward pass to route activations through newly injected parallel convolutional modules, introducing thousands to millions of new trainable parameters depending on the underlying model. Rather than a standalone method, Buffer Layers are designed to augment existing TTA methods. In contrast, SPeaR avoids this structural bloat entirely by inserting a minimal, low-dimensional steering primitive at an early stage boundary, adding only a few hundred parameters while leaving every pretrained weight unchanged. In addition, SPeaR operates as a fully standalone TTA instead of relying on or modifying other methods.

\begin{figure*}[t]
\centering
\includegraphics[width=0.78\textwidth]{"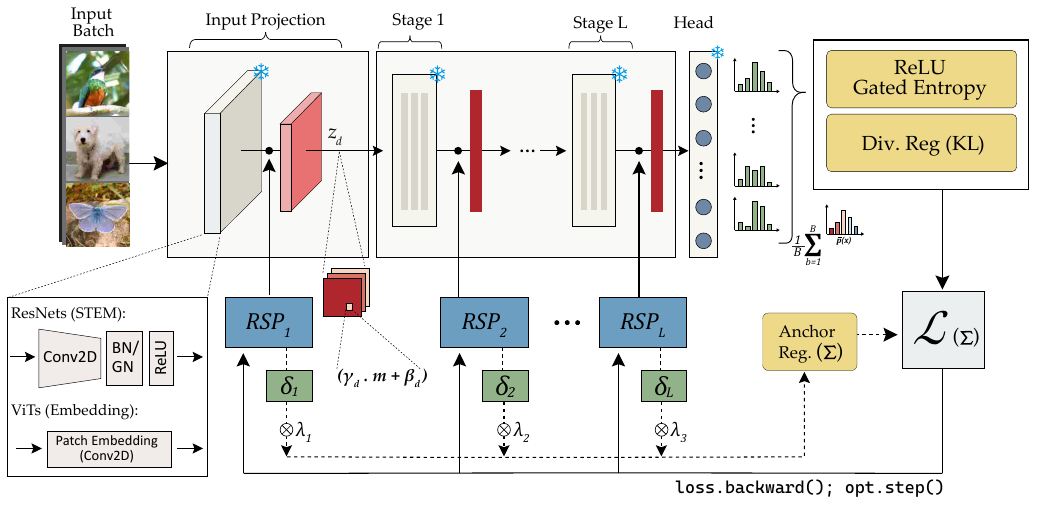"}
\caption{An overview of the SPeaR framework. RSPs are inserted after selected stages of a frozen classifier and optimized on the unlabeled test stream, using a soft-gated entropy term, a diversity regularizer, and a multi-depth anchor. 
}
\label{fig:graphical_abstract}
\end{figure*}

\section{Method}
\label{sec:method}

\subsection{Setup}
\label{sec:setup}

Let $f_\theta$ denote a network pretrained on a source distribution $\mathcal{D}_s$ with parameters $\theta$. At deployment, the model receives unlabeled data from a target distribution $\mathcal{D}_t \neq \mathcal{D}_s$ and must maintain performance without access to source data or target labels. Most conventional TTA methods update a subset of $\theta$. 
Our goal, however, is to keep $\theta$ entirely frozen by introducing a separate set of steering parameters $\phi$ that act on the representation as it propagates through the network, optimizing only $\phi$ during adaptation.

\noindent\textbf{Backbone Decomposition.} 
We denote the network as a stem, a sequence of $L$ stages, and a classifier head,
\begin{equation}
f_\theta = h \circ g_L \circ \cdots \circ g_1 \circ s.
\end{equation}
This decomposition is architecture-independent. In a convolutional network, $s$ is the initial convolutional block and $g_1, \dots, g_L$ are its residual stages. In a vision Transformer, $s$ is the patch-embedding layer and each $g_d$ is a contiguous group of Transformer encoder blocks.

\noindent\textbf{Stage Boundaries.} 
The decomposition defines $L+1$ boundaries, at which the intermediate representations are
\begin{equation}
z_0 = s(x), \qquad z_d = g_d(z_{d-1}), \quad d = 1, \dots, L,
\end{equation}
with the prediction given by $f_\theta(x) = h(z_L)$. Each $z_d$ is the output of a stage: 
in a convolutional network it is the output of a residual stage, and in a Transformer it is the token representation following the final block of an encoder group.

\noindent\textbf{Representation Structure.} 
Although their semantics differ, the representations at these boundaries share a common structure. Both can be represented as
$z_d \in R^{B \times C_d \times M_d}$,
where $C_d$ indexes channels in the convolutional case and embedding dimensions in the Transformer case, and $M_d$ indexes $M_d = H_d W_d$ positions or $M_d = N$ tokens. We refer to $C_d$ as the channel axis and $M_d$ as the position axis.

\subsection{Our Solution}
\label{sec:steering}

We propose a steering primitive, Representation Steering Primitive (RSP), that is parameterized along the channel axis and shared across the position axis, so that a single pair of parameters governs each channel regardless of how many positions it
spans. At boundary $d \in \mathcal{S}$, the RSP applies a per-channel scale and shift:
\begin{equation}
\text{RSP}_d(z_d)[b, c, m] = \gamma_d[c] \cdot z_d[b, c, m] + \beta_d[c],
\end{equation}
where $m$ is the activation in feature map $c$ in batch $b$, and $\gamma_d, \beta_d \in R^{C_d}$. 
Note that with $\gamma_d = \mathbf{1}$ and $\beta_d = \mathbf{0}$ the operation is identity, which gives both the initialization used throughout and a fixed reference point against which every subsequent update is measured (Section 3.4).

Let $\mathcal{S} \subseteq \{0, 1, \dots, L\}$ denote the set of boundaries at which an RSP is inserted, and let $\text{RSP}_d$ denote the adaptor at boundary $d$ with parameters $(\gamma_d, \beta_d)$. The adapted network computes
\begin{equation}
\begin{aligned}
\tilde{z}_0 &= \begin{cases} 
\text{RSP}_0(s(x)) & 0 \in \mathcal{S} \\ 
s(x) & \text{otherwise} 
\end{cases} \\[1ex]
\tilde{z}_d &= \begin{cases} 
\text{RSP}_d(g_d(\tilde{z}_{d-1})) & d \in \mathcal{S} \\ 
g_d(\tilde{z}_{d-1}) & \text{otherwise} 
\end{cases}
\end{aligned}
\end{equation}
and predicts $f_\phi(x) = h(\tilde{z}_L)$. The adapted parameter set is $\phi = \{(\gamma_d, \beta_d) : d \in \mathcal{S}\}$, of size
\begin{equation}
|\phi| = 2 \sum_{d \in \mathcal{S}} C_d.
\end{equation}
Every adaptor is initialized to the identity, so that $f_\phi = f_\theta$ before adaptation begins and the network starts at exactly the pretrained solution. Throughout adaptation $\theta$ remains frozen and only $\phi$ is updated. 


With no labels available, entropy minimization via the model's own average prediction is the natural choice to optimize $\phi$. However, directly minimizing entropy may introduce two well-known failure modes: the reinforcement of confident mistakes and class collapse. Therefore, we use gated entropy reduction containing two terms addressing the first, and a diversity term for the second. Additionally, we employ a hierarchical anchor regularization that limits how far the representation is steered to prevent catastrophic forgetting of the source model.

\noindent \textbf{Gated Entropy Reduction.}
The primary signal is the model's predictive entropy on each test batch. For an input $x_i$ we use the normalized entropy
\begin{equation}
\hat{H}(x_i) = \frac{-\sum_{k=1}^{K} p_k(x_i)\log p_k(x_i)}{\log K}, \label{eq:entropy}
\end{equation}
where $p(x_i) = \mathrm{softmax}(f_\theta(x_i))$ and $K$ is the number of classes. Dividing by $\log K$ bounds $\hat{H}$ to $[0,1]$ independent of $K$, so a single threshold $\tau$ carries the same impact across datasets with different label-space sizes. 
We compute the mean batch entropy $\overline{H} = \frac{1}{B}\sum_i \hat{H}(x_i)$, by averaging the prediction entropies across the current batch, which is used at two levels: a hard gate that determines whether adaptation of the current input batch is triggered and a soft gate that determines how far uncertainty minimization should take place.

\noindent \textbf{Hard Gate.} Before adapting a batch, we compare its mean entropy to $\tau$. If $\overline{H} < \tau$, the source model is already confident and we skip the batch, leaving the RSP unchanged. The hard gate produces no gradient; it only governs whether a batch is adapted at all, sparing confident batches from updates and the drift they may cause. Interestingly, this contrasts with EATA's sample selection strategy \cite{niu2022efficient}, which retains confident samples as the cleanest signal. 

Our findings, however, suggest that repeatedly minimizing the entropy of already confident predictions leads to severe over-optimization and confirmation bias for RSP. We present a detailed analysis in the Appendix G.

\noindent \textbf{Soft Gate.} 
At each step of the RSP optimization, we recompute the batch mean from the current predictions and apply a ReLU-gated penalty,
\begin{equation}
\mathcal{L}_{ent} = \mathrm{ReLU}(\overline{H} - \tau). \label{eq:loss_ent}
\end{equation}
As long as $\overline{H} > \tau$, the soft-gate stays active.
As $\overline{H}$ approaches $\tau$ the gradient shrinks and may eventually vanish once the batch is confident enough, after which only the other terms of the overall loss remain active. 

\noindent \textbf{Class-Diversity Regularization.}
Entropy minimization alone can cause class collapse, as the model may assign all inputs to a small number of classes \cite{mummadi2021test}. To prevent this, we push the mean prediction distribution over the batch to remain uniform across classes:
\begin{equation}
\small
    \mathcal{L}_\text{div} = D_{KL}(\overline{p} || \mathcal{U}),
\end{equation}
where $\overline{p} = \frac{1}{B} \sum_{i} p(x_{i})$ is the mean softmax prediction over the batch and $\mathcal{U} = \frac{1}{K}\mathbf{1}$ is the uniform distribution over $K$ classes. 

\noindent \textbf{Multi-depth Anchor Regularization.} 
When primitives at different depths of the network are used, those closer to the classification head have more direct influence over the model's output since a small perturbation at the last stage changes logits almost immediately, while the same perturbation at the stem is filtered through multiple nonlinear stages before reaching the classifier. As a result, deeper primitives are more sensitive to optimization and drift faster from their identity initialization, overshooting their optimal correction before shallower primitives can contribute meaningfully. 
To mitigate catastrophic forgetting due to model drift, especially in the continual TTA setup, we penalize each primitive's deviation from identity with a weight that increases with depth. We define an anchor loss as
\begin{equation}
    \mathcal{L}_{\text{anchor}} = \sum_{d=0}^{L} \lambda_d \left( \|\gamma_d - \mathbf{1}\|^2 + \|\beta_d\|^2 \right),
\end{equation}
where $\lambda_d$ is the weight for the primitive at depth $d$.
We use $\lambda_0 < \lambda_1 < \dots < \lambda_L$ so that deeper primitives, which have the most impact on predictions, are penalized more heavily for moving away from identity. Shallower primitives, whose corrections are naturally dampened by downstream processing, are given more freedom.

\noindent \textbf{Calculating Batch Statistics.}
For backbones that carry running statistics accumulated over the source data, we normalize using a weighted sum of the stored source statistics $\mathbf{s}_{\text{src}} = (\mu_{\text{src}}, \sigma^2_{\text{src}})$ and those of the current batch $\mathbf{s}_{\text{batch}}$, governed by a single coefficient $\alpha \in [0,1]$:
\begin{equation}
    \mathbf{s}_\alpha = (1-\alpha)\,\mathbf{s}_{\text{src}} + \alpha\,\mathbf{s}_{\text{batch}} .
\end{equation}
The stored statistics are never overwritten, no pretrained weights are modified, and this combination affects only the forward pass.

\noindent \textbf{Full Objective.}
The total objective $\mathcal{L}$ combines the three terms of Eqs. (7), (8), and (9), weighted by $\lambda_\text{ent}$, $\lambda_\text{div}$, and $\lambda_\text{anchor}$, respectively.
The total objective is defined as:
\begin{equation}
\label{eq:total_objective}
\mathcal{L} = \lambda_\text{ent}\mathcal{L}_\text{ent} + \lambda_\text{div}\mathcal{L}_\text{div} + \lambda_\text{anchor}\mathcal{L}_\text{anchor}.
\end{equation}
At each adaptation step, $\mathcal{L}$ is backpropagated through the RSP parameters $\gamma$ and $\beta$, which are updated using Adam while the backbone stays frozen. Following the standard online TTA protocol, each incoming batch is first evaluated using the model as adapted on previous batches. The batch is then used to update the model over a fixed number of adaptation steps, preparing it for the batches that follow.

\section{Experiment Setup}

\label{sec:experiments}

\noindent \textbf{Datasets.} We evaluate on three standard corruption benchmarks: CIFAR-10-C, CIFAR-100-C, and ImageNet-C \cite{hendrycks2019benchmarking}. Each applies 15 corruption types spanning four groups (noise, blur, weather, and digital). Following common practice in the TTA literature \cite{wang2020tent, niu2022efficient, wang2022continual}, we report results at the highest severity (level 5).

\noindent \textbf{Backbones.} 
We follow prior TTA papers and use ResNet-26 for CIFAR-10-C, ResNeXt-29 for CIFAR-100-C, and ResNet-50 and ResNet-50-GN for ImageNet-C. To further expand the experiments, we also use a ViT-B/16 \cite{rw2019timm, dosovitskiy2021an} on ImageNet-C. Details are presented in Appendix A.

\noindent \textbf{Evaluation.} We evaluate under two protocols: fully online on single out-of-distribution data, and continual adaptation. 
The stream ordering is fixed by a shared seed so that every method encounters an identical ordering.

\noindent \textbf{Baselines.} We compare against a range of established test-time adaptation methods operating under the same source-free constraints. For single corruption, we compare against TENT \cite{wang2020tent}, EATA \cite{niu2022efficient}, SAR \cite{niu2023towards}, ROID \cite{marsden2024universal}, DeYO \cite{lee2024entropy}, CMF \cite{lee2024continual}, and Buffer \cite{kim2025buffer} on three batch sizes. For continual TTA, we additionally include CoTTA \cite{wang2022continual} and RoTTA \cite{yuan2023robust}. All baselines are run on the identical data stream and evaluation protocol described above, with their published hyperparameters.

\section{Results}
\label{sec:morewithless}

\subsection{Single Shift Adaptation}

\begin{figure}[t]
    \centering
    \includegraphics[width=\linewidth]{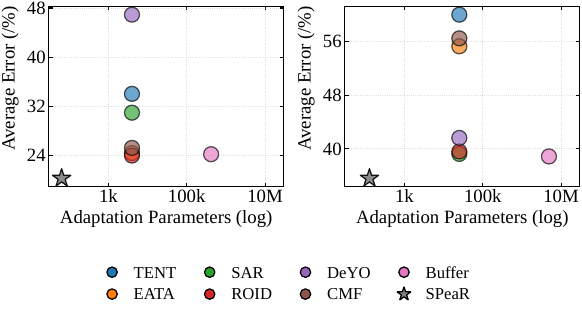}
\caption{State-of-the-art comparison of performance (percentage error, lower is better) versus the number of adapted parameters on CIFAR-10-C (ResNet-26) and CIFAR-100-C (ResNeXt-29), averaged across batch sizes. SPeaR achieves competitive performance while adapting orders of magnitude fewer parameters. Results for additional backbones are provided in Appendix F.}
    \label{fig:performancevsparameter}
\end{figure}

We report the results in Table \ref{tab:comprehensive_results} across CIFAR-10-C, CIFAR-100-C, and ImageNet-C under online adaptation at batch sizes 4, 16, and 256. 
Further per-corruption results are presented in Appendix C.
SPeaR's advantage is particularly pronounced at batch size 4, where several entropy-based methods degrade sharply while SPeaR remains stable, indicating substantially less dependence on large representative test batches. Following prior works, additional results on ImageNet-C with ResNet-50-GN and NFNet backbones are presented in Appendix D, where we observe significant gains using SPeaR. Finally, our experiments using ViT-B/16 on ImageNet-C with different batch sizes are presented in Table \ref{tab:alternate}, where our model achieves the best performance in two out of the three batch-size configurations. Per-corruption results are presented in Appendix E.
To obtain a high-level view of the effectiveness of our method against prior works, we provide a performance-parameter plot in Figure \ref{fig:performancevsparameter}. The analysis shows that despite adapting only 64 to 640 parameters, SPeaR consistently outperforms or achieves competitive results against state-of-the-art methods that update two to four orders of magnitude more parameters. 
Together, these results show that the same lightweight steering primitive remains effective across batch-normalized, group-normalized, normalization-free, and Transformer architectures without modifying the pretrained backbone.

\begin{table}[t]
\centering
\caption{Corruption error (\%) at severity 5 on CIFAR-10-C, CIFAR-100-C, and ImageNet-C, under the single-shift protocol across varying batch sizes.}
\label{tab:comprehensive_results}
\setlength{\tabcolsep}{3pt}
\resizebox{\columnwidth}{!}{%
\begin{tabular}{l ccc ccc ccc}
\toprule
& \multicolumn{3}{c}{\textbf{CIFAR-10-C}} & \multicolumn{3}{c}{\textbf{CIFAR-100-C}} & \multicolumn{3}{c}{\textbf{ImageNet-C}} \\
\cmidrule(lr){2-4} \cmidrule(lr){5-7} \cmidrule(lr){8-10}
Method & 4 & 16 & 256 & 4 & 16 & 256 & 4 & 16 & 256 \\
\midrule
Source & 43.13 & 43.13 & 43.13 & 46.45 & 46.45 & 46.45 & 80.81 & 80.81 & 80.81 \\
TENT & 58.29 & 24.33 & 19.45 & 94.64 & 53.45 & 31.85 & 78.48 & 63.45 & 61.21 \\
EATA & 33.18 & 22.13 & 17.79 & 90.15 & 44.21 & 31.45 & 81.06 & 62.66 & 58.16 \\
SAR & 45.38 & 25.19 & 22.28 & 49.05 & \textbf{34.12} & 34.59 & 81.38 & 65.87 & 60.67 \\
ROID & 32.14 & 21.20 & 18.57 & 52.83 & 34.87 & 31.15 & 77.89 & \textbf{59.75} & 56.16 \\
DeYO & 86.32 & 35.20 & 19.25 & 54.35 & 35.43 & 35.12 & 81.42 & 65.08 & 60.62 \\
CMF & 33.69 & 22.32 & 19.65 & 93.63 & 43.89 & 31.82 & 79.28 & 62.25 & 61.73 \\
Buffer & 29.65 & 20.95 & 21.97 & 48.25 & 34.93 & 33.44 & 83.28 & 71.96 & 60.84 \\
SPeaR & \textbf{24.70} & \textbf{19.17} & \textbf{16.93} & \textbf{41.93} & 34.25 & \textbf{30.69} & \textbf{65.90} & 60.91 & \textbf{54.80} \\
\bottomrule
\end{tabular}
}
\end{table}

\begin{table}[t]
\centering
\caption{Corruption error (\%) at severity 5 on ImageNet-C (ViT-B/16) under the single shift protocol across varying batch sizes.}
\label{tab:alternate}
\setlength{\tabcolsep}{6pt}
\resizebox{0.5\columnwidth}{!}{%
\begin{tabular}{l ccc}
\toprule
Method & 4 & 16 & 256 \\
\midrule
Source & 48.37 & 48.37 & 48.37 \\
SAR    & \textbf{34.70} & 33.75 & 36.66 \\
ROID   & 39.73 & 31.99 & 36.07 \\
DeYO   & 86.11 & 40.06 & 37.10 \\
SPeaR  & 36.16 & \textbf{31.13} & \textbf{32.10} \\
\bottomrule
\end{tabular}
}
\end{table}

\begin{table*}[t]
\centering
\caption{Performance on CIFAR-10-C, CIFAR-100-C, and ImageNet-C under continual  adaptation, following CoTTA's standard setup at maximum severity (5).}
\label{tab:continual}
\setlength{\tabcolsep}{5pt}
\resizebox{\textwidth}{!}{%
\renewcommand{\arraystretch}{0.9} 
\begin{tabular}{cl ccc cccc cccc cccc >{\columncolor{summaryblue}}c}
\toprule
& & \multicolumn{3}{c}{Noise} 
& \multicolumn{4}{c}{Blur} 
& \multicolumn{4}{c}{Weather} 
& \multicolumn{4}{c}{Digital} 
& \\
\cmidrule(lr){3-5} \cmidrule(lr){6-9} \cmidrule(lr){10-13} \cmidrule(lr){14-17}
Dataset & Method 
& GN & SN & IN  
& DB  & GB  & MB  & ZB  
& Snow & Frost & Fog & Bright 
& Cont & Elastic  & Pixel & JPEG
& Mean \\
\midrule

\multirow{6}{*}{\rotatebox[origin=c]{-270}{\scriptsize \begin{tabular}{@{}c@{}}\textbf{CIFAR-10-C} \end{tabular}}} 
& CoTTA & 23.84 & 20.50 & \textbf{24.98} & \textbf{12.54} & \textbf{25.31} & \textbf{14.20} & 13.22 & 16.77 & 15.10 & 14.15 & 9.97 & 15.42 & \textbf{18.97} & 15.80 & \textbf{17.96} & 17.26 \\
& RoTTA & 40.55 & 38.49 & 47.86 & 15.83 & 41.96 & 17.73 & 16.30 & 24.53 & 32.96 & 17.02 & 13.53 & 46.15 & 29.40 & 28.38 & 34.83 & 29.70 \\
& DeYO & 25.03 & 22.69 & 31.01 & 19.14 & 35.27 & 23.33 & 19.74 & 20.65 & 21.64 & 21.95 & 15.24 & 16.12 & 29.58 & 26.23 & 34.18 & 24.12 \\
& CMF & 26.02 & 20.77 & 29.86 & 13.53 & 30.62 & 15.40 & 12.51 & 16.91 & 16.49 & 14.89 & 8.78 & 11.96 & 22.23 & 17.14 & 23.00 & 18.67 \\
& Buffer & 27.02 & \textbf{18.54} & 25.37 & 13.81 & 32.98 & 18.89 & 14.79 & 18.30 & 17.98 & 15.21 & 10.65 & 14.26 & 22.73 & 19.44 & 25.61 & 19.71 \\
& SPeaR & \textbf{22.23} & 19.12 & 29.72 & 16.07 & 29.79 & 15.37 & \textbf{11.96} & \textbf{14.10} & \textbf{14.62} & \textbf{11.57} & \textbf{8.09} & \textbf{10.09} & 21.12 & \textbf{15.12} & 19.82 & \textbf{17.25} \\

\midrule

\multirow{6}{*}{\rotatebox[origin=c]{-270}{\scriptsize \begin{tabular}{@{}c@{}}\textbf{CIFAR-100-C} \end{tabular}}} 
& CoTTA & 40.14 & 38.72 & 40.05 & 30.38 & 40.29 & 32.20 & 30.64 & 36.86 & 34.21 & 45.84 & 31.06 & 35.17 & 39.12 & 34.28 & 40.44 & 36.63 \\
& RoTTA & 49.84 & 46.25 & 47.11 & 31.84 & 44.62 & 31.30 & 28.16 & 34.66 & 33.84 & 42.56 & 26.35 & 34.00 & 36.08 & 32.76 & 38.39 & 37.18 \\
& DeYO & 41.52 & 36.13 & 37.57 & 26.85 & \textbf{38.15} & 28.90 & \textbf{26.54} & \textbf{32.37} & \textbf{31.23} & 36.26 & 25.83 & \textbf{28.23} & \textbf{33.09} & 29.69 & 37.49 & 32.66 \\
& CMF & 38.97 & 41.89 & 51.79 & 54.27 & 78.31 & 81.71 & 85.70 & 94.20 & 96.66 & 97.18 & 97.85 & 97.83 & 97.69 & 97.78 & 97.86 & 80.65 \\
& Buffer & \textbf{37.71} & \textbf{33.76} & 36.60 & 29.13 & 38.43 & 32.01 & 29.04 & 34.15 & 33.39 & 37.33 & 28.21 & 31.38 & 34.97 & 30.69 & 38.76 & 33.70 \\
& SPeaR & 37.96 & 36.97 & \textbf{32.58} & \textbf{26.47} & 38.94 & \textbf{28.72} & 26.78 & 32.92 & 32.73 & \textbf{35.97} & \textbf{25.29} & 28.91 & 35.43 & \textbf{29.30} & \textbf{37.31} & \textbf{32.42} \\

\midrule

\multirow{6}{*}{\rotatebox[origin=c]{-270}{\scriptsize \begin{tabular}{@{}c@{}}\textbf{ImageNet-C} \end{tabular}}} 
& CoTTA & 82.99 & 79.04 & 76.61 & 80.98 & 77.67 & 70.20 & 60.92 & 62.66 & 62.77 & 52.94 & 36.57 & 75.77 & 53.18 & 47.33 & 53.46 & 64.87 \\
& RoTTA &  78.95 & 70.37 & 65.10 & 77.67 & 72.00 & 71.60 & 65.90 & 71.05 & 69.72 & 67.87 & 53.74 & 74.02 & 64.34 & 60.43 & 62.74 & 68.37  \\
& DeYO & 76.39 & 65.60 & \textbf{63.87} & \textbf{73.81} & \textbf{68.67} & \textbf{60.48} & \textbf{52.10} & 58.76 & \textbf{58.05} & 47.26 & 35.39 & \textbf{59.49} & \textbf{46.36} & 42.57 & 46.79 & 57.04 \\
& CMF & 76.88 & 68.80 & 68.00 & 75.04 & 72.32 & 63.24 & 52.37 & 59.02 & 59.79 & 46.27 & 33.84 & 69.86 & 48.06 & 43.71 & 49.43 & 59.11 \\
& Buffer & 83.74 & 99.49 & 99.71 & 99.80 & 99.74 & 99.79 & 99.79 & 99.79 & 99.82 & 99.85 & 99.80 & 99.85 & 99.93 & 99.93 & 99.88 & 98.73 \\

& SPeaR & \textbf{67.11} & \textbf{63.92} & 64.18 & 75.77 & 71.23 & 64.72 & 58.26 & \textbf{54.76} & 58.50 & 43.64 & 33.13 & 61.63 & 49.15 & \textbf{41.40} & \textbf{45.91} & \textbf{56.89} \\

\bottomrule
\end{tabular}
}
\end{table*}

\subsection{Continual Adaptation}

\noindent \textbf{Performance.} We evaluate SPeaR under continual adaptation, where the model adapts sequentially across all 15 corruptions without resetting between shifts, following the protocol in CoTTA \cite{wang2022continual}. The steering primitive configuration is kept fixed throughout the experiment. Results for prior methods are reproduced using their publicly available implementations. For consistency, all methods including SPeaR use the same hyperparameter settings as in the single-shift TTA experiments, without additional tuning for the continual setting. As shown in Table \ref{tab:continual}, SPeaR consistently achieves competitive or superior performance compared with existing methods, demonstrating its effectiveness under continuously evolving distribution shifts.

\begin{figure}[t]
    \centering
    \includegraphics[width=1\linewidth]{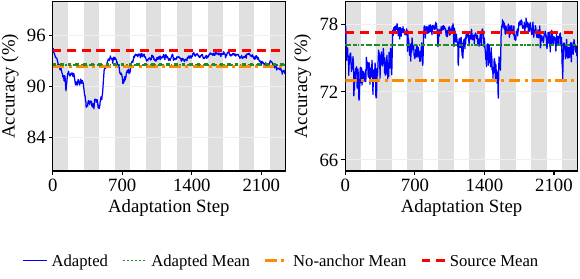}
    \caption{Accuracy on non-corrupted samples during continual adaptation on CIFAR-10-C (left) and CIFAR-100-C (right). SPeaR's original accuracy on non-corrupted samples (with and without the anchor) is shown against the source model's original accuracy; alternating grey and white bands mark the 15 corruptions in sequence.}
    \label{fig:continual}
\end{figure}

\noindent \textbf{Source Preservation.}
Continual adaptation introduces the risk of catastrophic forgetting, where adaptation to successive distribution shifts degrades knowledge acquired during source training. Figure \ref{fig:continual} evaluates SPeaR's ability to preserve source knowledge during continual adaptation. On CIFAR-10-C, SPeaR maintains a mean accuracy of 92.52\%, approximately 2 percentage points below the source model's performance on the original (non-corrupted) data. On CIFAR-100-C, non-corrupted accuracy averages 76.15\%, compared to 77.29\% for the source model. More importantly, performance remains stable throughout adaptation, with only minor variation over the incoming test stream (+1.24 and +0.92 points per 1K batches on CIFAR-10-C and CIFAR-100-C, respectively), indicating no progressive forgetting. Figure \ref{fig:continual} also reports continual adaptation without the anchoring objective. Removing the anchor has little effect on CIFAR-10-C, reducing average accuracy by only 0.15 points, but leads to a substantially larger drop of approximately 3 points on CIFAR-100-C. Overall, SPeaR preserves source-domain performance throughout continual adaptation, while the anchor becomes increasingly important in the more challenging CIFAR-100-C setting.

\subsection{Ablation Studies}
We conduct ablation studies to isolate the contribution of each component of the RSP objective. We place RSPs at every stage boundary and adapt them throughout the experiments. As shown in Table \ref{tab:ablation}, the complete objective achieves the best performance on both datasets. We first remove the batch statistics blending coefficient $\alpha$, which increases corruption error from 20.63\% to 23.99\% on CIFAR-10-C and from 34.15\% to 40.79\% on CIFAR-100-C, demonstrating the benefit of retaining source statistics during adaptation. Starting from this configuration, removing gated entropy causes a much larger degradation, increasing corruption error to 39.18\% and 74.41\%, confirming that entropy provides the primary adaptation signal. Removing diversity regularization also degrades performance, increasing corruption error to 58.72\% and 68.00\%, consistent with its role in preventing class collapse. In contrast, removing the anchor results in a smaller but consistent degradation, from 23.99\% to 24.76\% on CIFAR-10-C and from 40.79\% to 41.45\% on CIFAR-100-C. 
Note that the anchor is not designed to drive adaptation, but to constrain the primitives toward their identity initialization and limit source forgetting. Accordingly, its removal has a modest effect on corruption error but consistently increases clean error, from 20.19\% to 21.04\% on CIFAR-10-C and from 33.07\% to 33.87\% on CIFAR-100-C. Its effect is more pronounced when combined with the prediction-driven objectives, where it regularizes their adaptation signal and limits excessive representation drift.

\begin{table}[t]
\centering
\caption{Ablation study on CIFAR-10-C and CIFAR-100-C. Error (\%) is measured at severity 5.}
\label{tab:ablation}
\footnotesize
\setlength{\tabcolsep}{2.8pt}
\begin{tabular}{@{} c ccc cc cc @{}}
\toprule
& \multicolumn{3}{c}{Objectives} & \multicolumn{2}{c}{CIFAR-10-C} & \multicolumn{2}{c}{CIFAR-100-C} \\
\cmidrule(lr){2-4} \cmidrule(lr){5-6} \cmidrule(lr){7-8}
$\alpha$ & $\mathcal{L}_{ent}$ & $\mathcal{L}_{div}$ & $\mathcal{L}_{anchor}$ & Corrupt. & Clean & Corrupt. & Clean \\
\midrule

\checkmark & \checkmark & \checkmark & \checkmark & 20.63 & 9.15 & 34.15 & 27.53 \\
\midrule
$\times$ & \checkmark & \checkmark & \checkmark & 23.99 & 20.19 & 40.79 & 33.07 \\
$\times$ & $\times$ & \checkmark & \checkmark & 39.18 & 31.84 & 74.41 & 74.86 \\
$\times$ & \checkmark & $\times$ & \checkmark & 58.72 & 18.63 & 68.00 & 48.99 \\
$\times$ & \checkmark & \checkmark & $\times$ & 24.76 & 21.04 & 41.45 & 33.87 \\
\midrule
$\times$ & \checkmark & $\times$ & $\times$ & 61.56 & 17.36 & 68.36 & 49.38 \\
$\times$ & $\times$ & \checkmark & $\times$ & 41.59 & 36.69 & 79.79 & 82.74 \\
$\times$ & $\times$ & $\times$ & \checkmark & 43.12 & 4.91 & 46.45 & 54.12 \\
\bottomrule
\end{tabular}
\end{table}

\begin{figure*}[t]
\centering
\includegraphics[width=\textwidth]{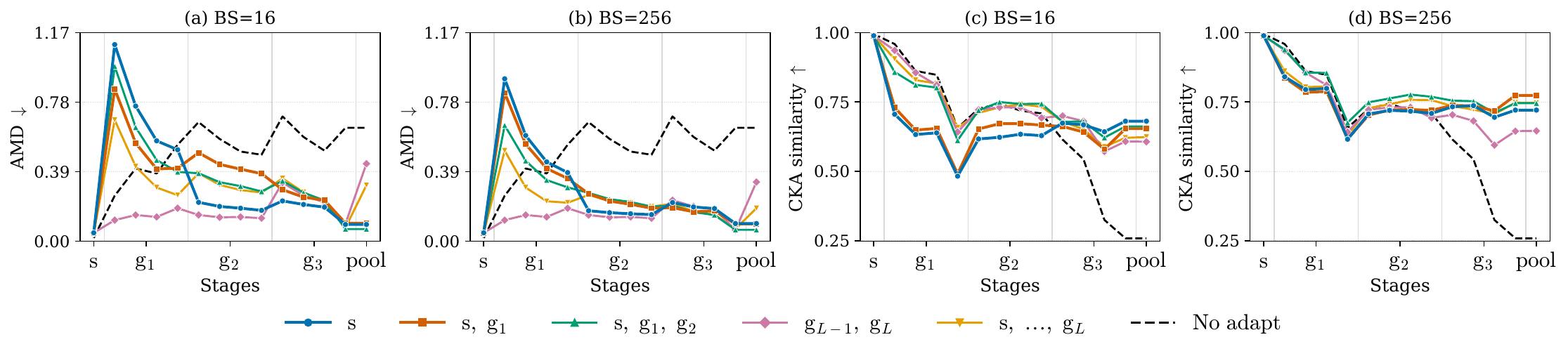}
\caption{AMD ($\downarrow$) and CKA ($\uparrow$) relative to the source reference across the encoder blocks (stem $\rightarrow$ classifier) for five RSP placement configurations and a no-adaptation baseline on CIFAR-10-C (severity 5) at batch sizes 16 and 256. Probes placed before each stage boundary track representation dynamics throughout the network. Early steering initially perturbs intermediate representations but produces the closest alignment of the final pre-classifier representation, whereas deeper placement leaves substantially larger residual mismatch.}
\label{fig:drift_cka}
\end{figure*}

\subsection{Configuration Analysis}

\noindent \textbf{Feature-level Analysis.} 
We investigate how the placement of RSPs influences representation dynamics throughout the network. Using ResNet-26 on CIFAR-10-C, we compare the representations produced by corrupted inputs after adaptation with the corresponding representations of the frozen source model on clean inputs. We quantify their similarity using Activation Mean Deviation (AMD) and Centered Kernel Alignment (CKA) \cite{kornblith2019similarity}. AMD measures deviations in channel-wise activation statistics, whereas CKA quantifies the similarity of representation geometry through pairwise sample relationships. Figure~\ref{fig:drift_cka} reports AMD (left) and CKA (right) across different batch sizes. We find that placing the RSP at early network stages produces the smallest activation deviation and the highest CKA similarity at the final-stage representation, whereas deeper placements leave the final features substantially farther from the reference. Interestingly, this recovery does not arise from preserving intermediate representations. Early RSPs initially induce noticeable deviations from the reference features, yet these deviations are progressively transformed by subsequent layers, yielding a final representation that closely matches the source model. This suggests that an RSP does not need to restore features locally at its insertion point. Instead, it steers the representation onto a trajectory that compensates for the effects of distribution shift as they accumulate through the network. Among the early configurations, the preferred placement depends on batch size: a single RSP at the stem yields the closest alignment at small batch sizes, whereas adding a second RSP after the first stage produces the closest alignment at larger batch sizes.

\begin{figure}[t]

\centering
\includegraphics[width=1.0\columnwidth]{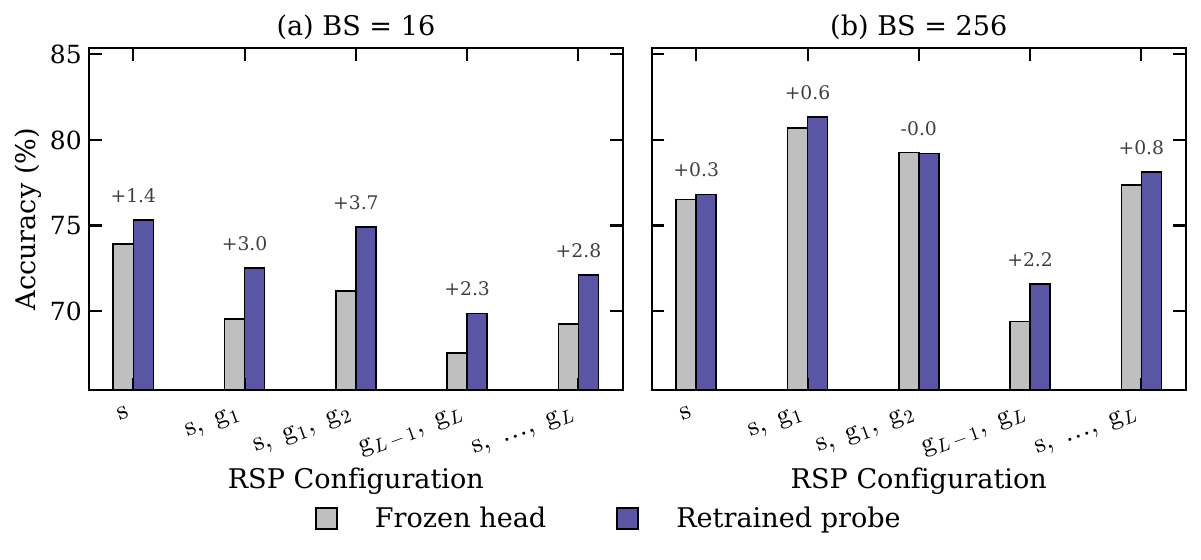}
\caption{Linear-probe analysis on CIFAR-10-C for the five RSP placement configurations at batch sizes 16 (left) and 256 (right). 
Early RSP placement consistently yields the highest accuracy and the smallest probe gaps, with the preferred early configuration depending on batch size.}
\label{fig:sep_bars}
\end{figure}

\noindent \textbf{Linear Probing.} 
To further validate these findings, we evaluate the adapted representations using both the adapted model and a linear probe retrained on the same model \cite{chen2020simple, alain2018understanding}. The results of this experiment for different RSP configurations are presented in Figure~\ref{fig:sep_bars}.
Our analysis leads to two observations. First, early RSP placement produces more discriminative representations. Across both batch sizes, early configurations consistently achieve higher frozen-classifier accuracy and smaller probe gaps than the deeper configurations. In contrast, \textit{deep} placement performs worst, yielding the lowest accuracy and among the largest probe gaps (2.3 points at batch size 16 and 2.2 points at batch size 256), indicating representations that are less aligned with the source classifier. Second, the preferred early configuration depends on batch size. At batch size 16, using a single RSP achieves both the highest accuracy and the smallest probe gap (1.4 points), while at batch size 256, two early RSPs perform best, matching the accuracy of the retrained probe while outperforming the remaining configurations. This suggests that the additional steering capacity of a second RSP is beneficial when larger batches provide a more reliable adaptation signal, whereas a single RSP remains preferable in the low-batch regime.

\begin{figure}[t]

\centering
\includegraphics[width=\columnwidth]{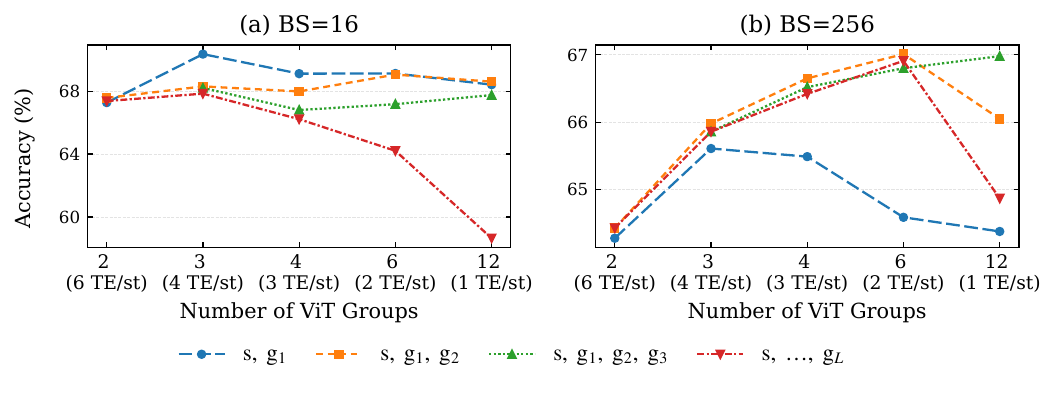}
\caption{Performance of different RSP placement configurations across ViT-B/16 stage partitionings on ImageNet-C (severity 5), evaluated at batch sizes 16 (left) and 256 (right). 
Larger batches benefit from finer partitionings, whereas smaller batches favor coarser partitionings.}
\label{fig:vit}
\end{figure}

\noindent \textbf{Steering Boundaries in Transformers.} 
ViT-B/16 contains twelve encoder blocks. We partition them into 2, 3, 4, 6, or 12 sequential groups, placing an RSP at the boundary of each selected group (Figure~\ref{fig:vit}). Thus, moving from left to right in Figure~\ref{fig:vit} progressively increases the number of possible RSP locations. For each grouping, we evaluate all placement configurations. 
Two consistent patterns emerge. At larger batch sizes, where each update is estimated from more samples, finer groupings with more active RSPs perform better. At smaller batch sizes, where the adaptation signal is noisier, coarser groupings with fewer RSPs are preferable. This mirrors the trend observed for residual networks. Notably, increasing the number of adapted parameters does not necessarily improve performance.

\section{Summary}

We introduced SPeaR, a test-time adaptation framework that keeps the pretrained model frozen and instead adapts the representations flowing through it. By inserting lightweight Representation Steering Primitives at early stage boundaries, SPeaR performs effective source-free adaptation without supervised warm-up. Across CIFAR-10-C, CIFAR-100-C, and ImageNet-C, SPeaR consistently matched or outperformed methods that adapt orders of magnitude more parameters, remained robust across a wide range of batch sizes, and generalized across both convolutional and Transformer backbones. Under continual adaptation, it also preserved clean-domain performance over long sequences of distribution shifts. Our results suggest that directly steering intermediate representations at stage boundaries, rather than adapting model parameters, is a simple, efficient, and broadly applicable paradigm for test-time adaptation.

\bibliography{aaai2026}


\clearpage

\appendix

\begin{center}
    \Large\bfseries Appendix
\end{center}

\section{Backbone Architectures and Model Details}
\label{sec:backbones}

We evaluate SPeaR across six pretrained backbones chosen to span the
design axes most relevant to test-time adaptation: multiple variants of ResNets, together with a vision Transformer. Table~\ref{tab:models} lists each backbone with the benchmark it is used on, its normalization scheme, and its parameter count. Every backbone is evaluated from the same pretrained checkpoint across all baselines, and is kept entirely frozen during adaptation. SPeaR introduces only the Representation Steering Primitives (RSPs) of Section~3, and no pretrained weight is modified in
any experiment.

\paragraph{CIFAR-10-C (ResNet-26).}

We train the ResNet-26 source model on clean CIFAR-10, following standard practice~\cite{sun2020test,wang2020tent,he2016resnet}.

\paragraph{CIFAR-100-C (ResNeXt-29).}

We use the AugMix-trained ResNeXt-29 ($4{\times}32$d) checkpoint~\cite{xie2017aggregated,hendrycks2019augmix} distributed by RobustBench~\cite{croce2020robustbench}. AugMix is applied only during source training; no augmentation is used at test time.
 
\paragraph{ImageNet-C (ResNet-50, ResNet-50-GN, NFNet, ViT).}

For the ImageNet-C results reported in Table~1, we use the standard BN ResNet-50 from \texttt{torchvision}. The additional experiments use a group-normalized ResNetV2-50d also known as ResNet-50-GN and the normalization-free NFNet-F0~\cite{brock2021high}, a model with no normalization layer. The Transformer results use ViT-B/16~\cite{dosovitskiy2021an}, pretrained on ImageNet-21k and fine-tuned on ImageNet-1k with AugReg~\cite{steiner2021train}. All ImageNet-C inputs are evaluated at $224\times224$.

\begin{table}[h]
\centering
\caption{Pretrained backbones used to evaluate SPeaR. Parameter counts are for the frozen backbone only; RSPs add at most a few thousand parameters. BN, GN, LN, and NA denote batch, group, layer, and no normalization, respectively.}
\small
\setlength{\tabcolsep}{4pt}
\begin{tabular}{llcc}
\toprule
Backbone & Benchmark & Norm. & \# Params \\
\midrule
ResNet-26 & CIFAR-10-C & BN & $1.47$\,M \\
ResNeXt-29 ($4{\times}32$d) & CIFAR-100-C & BN & $6.90$\,M \\
ResNet-50 & ImageNet-C & BN & $25.56$\,M \\
ResNet-50-GN & ImageNet-C & GN & $25.57$\,M \\
ViT-B/16 & ImageNet-C & LN & $86.57$\,M \\
NFNet-F0 & ImageNet-C & NA & $71.49$\,M \\
\bottomrule
\end{tabular}
\label{tab:models}
\end{table}

\begin{table}[h]
\centering
\caption{Channel width $C_d$ at each stage boundary ($C_0$ is the stem
output). These widths determine the insertion points and sizes of the RSPs: a primitive
at boundary $d$ has $2C_d$ parameters.}
\small
\setlength{\tabcolsep}{5pt}
\begin{tabular}{lcccccc}
\toprule
Backbone & $C_0$ & $C_1$ & $C_2$ & $C_3$ & $C_4$ & $L$ \\
\midrule
ResNet-26     & 32 & 32  & 64  & 128  & ---  & 3 \\
ResNeXt-29    & 64 & 256 & 512 & 1024 & ---  & 3 \\
ResNet-50     & 64 & 256 & 512 & 1024 & 2048 & 4 \\
ResNet-50-GN  & 64 & 256 & 512 & 1024 & 2048 & 4 \\
NFNet-F0      & 64 & 256 & 512 & 1536 & 1536 & 4 \\
\bottomrule
\end{tabular}

\label{tab:app_stages}
\end{table}

\section{Batch-Statistics Mixing Coefficient ($\alpha$)}

\begin{figure}[t]
    \centering
    \includegraphics[width=1\linewidth]{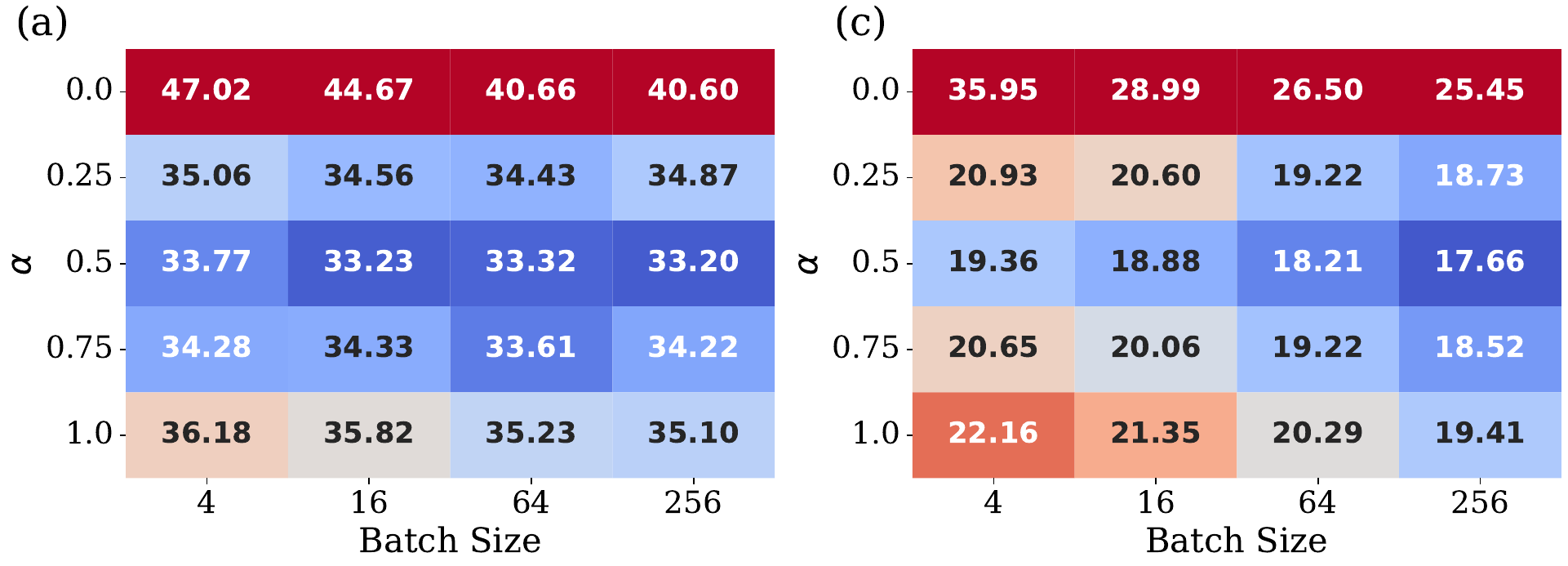}
    \caption{Corruption error for different values of $\alpha$ and test-batch size, on (a) CIFAR-100-C and (b) CIFAR-10-C. The total number of updates is held fixed across batch sizes. On both datasets $\alpha = 0.5$ is optimal at every batch size, and the optimum does not shift with batch size.}
    \label{fig:alpha}
\end{figure}

Methods that adapt batch-normalization affine parameters normalize using the statistics of the current test batch, replacing the source statistics entirely. This makes them dependent on those statistics being reliable, and inapplicable where a batch provides none. SPeaR takes neither position: rather than discarding the source statistics or relying wholly on the batch, it combines them through the coefficient $\alpha$ of Section~3, and we determine the value that serves it best empirically. We sweep $\alpha \in \{0, 0.25, 0.5, 0.75, 1\}$ against test-batch sizes $\{4, 16, 64, 256\}$ on CIFAR-10-C and CIFAR-100-C, holding the total number of updates fixed across batch sizes so that only the reliability of the batch statistics varies. Figure~\ref{fig:alpha} reports the resulting corruption error. The optimum is consistent. On both datasets, $\alpha = 0.5$ gives the lowest error at every batch size, with the minimum remaining in the same row of each heatmap rather than drifting as batch size changes. The best correction draws on both, and how much to trust the incoming batch is stable across the range of batch sizes tested. We use $\alpha = 0.5$ on both CIFAR benchmarks throughout.

\section{Complete Benchmark Results}
\label{sec:full_result_1}

Tables~\ref{tab:c10c}, \ref{tab:c100c}, and~\ref{tab:inc} expand Table~1 of the main paper into per-corruption results at severity~5, covering every corruption type and method at batch sizes $4$, $16$, and $256$. Across these backbones, SPeaR is particularly effective in the small-batch regime. Here, several methods that adapt model parameters, particularly those based on BN, perform worse than the unadapted source model, whereas SPeaR adapts consistently from the same batches. This suggests that representation steering depends less on large, statistically representative batches than approaches based on recalibrating normalization statistics or minimizing batch entropy. As the batch size increases, the performance gap narrows, yet SPeaR continues to achieve the best or near-best mean error across all three benchmarks.


\begin{table*}[t]
\centering
\caption{Per-corruption error (\%) at severity~5 on CIFAR-10-C, batch sizes $4$, $16$, $256$.}
\label{tab:c10c}
\setlength{\tabcolsep}{5pt}
\resizebox{\textwidth}{!}{%
\renewcommand{\arraystretch}{0.9} 
\begin{tabular}{cl ccc cccc cccc cccc >{\columncolor{summaryblue}}c}
\toprule
& & \multicolumn{3}{c}{Noise} 
& \multicolumn{4}{c}{Blur} 
& \multicolumn{4}{c}{Weather} 
& \multicolumn{4}{c}{Digital} 
& \\
\cmidrule(lr){3-5} \cmidrule(lr){6-9} \cmidrule(lr){10-13} \cmidrule(lr){14-17}
BS & Method 
& GN & SN & IN  
& DB  & GB  & MB  & ZB  
& Snow & Frost & Fog & Bright 
& Cont & Elastic  & Pixel & JPEG
& Mean \\
\midrule



\multirow{6}{*}{%
  \raisebox{-10.ex}{%
    \rotatebox[origin=c]{0}{\small 4}%
  }%
}





& Source & 69.58 & 63.89 & 72.98 & 40.85 & 50.90 & 34.63 & 37.27 & 23.87 & 37.75 & 31.43 & 8.96 & 71.06 & 25.11 & 50.80 & 27.86 & 43.13 \\
\midrule
& TENT & 64.95 & 66.86 & 73.33 & 54.13 & 73.53 & 47.25 & 55.60 & 52.91 & 46.19 & 51.72 & 45.57 & 58.67 & 66.15 & 54.48 & 63.00 & 58.29 \\

& EATA & 39.20 & 36.63 & 48.26 & 26.70 & 52.54 & 27.74 & 27.92 & 29.40 & 30.26 & 26.70 & 21.44 & 23.70 & 38.77 & 30.41 & 38.07 & 33.18 \\

& SAR & 57.97 & 54.33 & 61.63 & 39.71 & 61.74 & 39.16 & 35.91 & 37.30 & 45.88 & 34.25 & 25.25 & 31.24 & 54.84 & 49.00 & 52.47 & 45.38 \\

& ROID & 38.22 & 36.89 & 45.72 & 25.82 & 45.25 & 26.75 & 26.74 & 29.43 & 30.53 & 25.63 & 21.69 & 24.00 & 36.04 & 31.95 & 37.39 & 32.14 \\

& DeYO & 88.32 & 88.86 & 86.40 & 90.61 & 84.86 & 87.55 & 84.78 & 89.40 & 86.31 & 82.30 & 81.65 & 84.22 & 87.20 & 87.58 & 84.79 & 86.32 \\

& CMF & 40.31 & 37.53 & 49.37 & 27.21 & 51.11 & 26.71 & 27.45 & 30.58 & 30.84 & 27.22 & 22.55 & 25.66 & 36.88 & 33.29 & 38.60 & 33.69 \\

& Buffer & 34.99 & 32.75 & 42.08 & 24.32 & 42.60 & 25.15 & 24.79 & 26.15 & 27.51 & 24.59 & 19.75 & 22.48 & 34.92 & 27.83 & 34.78 & 29.65 \\

& SPeaR & \textbf{30.29} & \textbf{28.15} & \textbf{39.98} & \textbf{18.59} & \textbf{40.24} & \textbf{19.03} & \textbf{20.09} & \textbf{21.10} & \textbf{22.50} & \textbf{19.02} & \textbf{12.96} & \textbf{17.84} & \textbf{29.62} & \textbf{22.39} & \textbf{28.63} & \textbf{24.70} \\

\midrule

\multirow{6}{*}{%
  \raisebox{-5.ex}{%
    \rotatebox[origin=c]{0}{\small 16}%
  }%
}





& TENT & 30.49 & 31.19 & 41.20 & 17.00 & 36.10 & 19.02 & 16.85 & 20.74 & 20.79 & 19.62 & 13.19 & 21.12 & 29.02 & 21.10 & 27.47 & 24.33 \\

& EATA  & 27.58 & 25.69 & 35.42 & 15.53 & 38.98 & 17.58 & 16.07 & 19.10 & 20.10 & 15.24 & 12.09 & 13.77 & 26.95 & 20.38 & 27.51 & 22.13 \\

& SAR & 33.92 & 32.05 & 39.60 & 16.43 & 39.50 & 18.28 & 17.58 & 23.49 & 24.40 & 18.00 & 13.26 & 17.06 & 27.58 & 25.29 & 31.42 & 25.19 \\

& ROID & 26.93 & 24.36 & \textbf{33.08} & 15.44 & \textbf{32.46} & 17.13 & 15.37 & 18.96 & 19.91 & 16.24 & 12.47 & 15.97 & 24.49 & 19.96 & 25.17 & 21.20 \\

& DeYO & 49.65 & 38.62 & 56.24 & 24.62 & 59.92 & 27.68 & 24.73 & 28.52 & 27.44 & 21.52 & 17.65 & 19.22 & 57.50 & 36.86 & 37.83 & 35.20 \\

& CMF & 28.56 & 26.57 & 36.27 & 15.96 & 34.75 & 16.38 & 15.50 & 19.26 & 19.49 & 16.63 & 11.79 & 19.64 & 26.57 & 20.08 & 27.30 & 22.32 \\

& Buffer & 26.14 & 24.81 & 34.79 & 15.02 & 32.90 & 15.89 & 15.52 & 17.69 & 19.17 & 14.69 & 11.23 & 12.96 & 27.03 & 19.70 & 26.69 & 20.95 \\

& SPeaR & \textbf{24.40} & \textbf{22.67} & 33.94 & \textbf{13.23} & 33.47 & \textbf{15.05} & \textbf{14.06} & \textbf{16.25} & \textbf{17.76} & \textbf{13.86} & \textbf{8.66} & \textbf{11.39} & \textbf{23.40} & \textbf{17.12} & \textbf{22.34} & \textbf{19.17} \\


\midrule

\multirow{6}{*}{%
  \raisebox{-5.ex}{%
    \rotatebox[origin=c]{0}{\small 256}%
  }%
}








& TENT & 26.90 & 24.54 & 34.02 & 12.01 & 32.31 & 13.58 & 12.11 & 16.91 & 17.74 & 13.30 & 9.01 & 13.98 & 22.24 & 18.64 & 24.50 & 19.45 \\

& EATA & 23.94 & 21.56 & 31.11 & 11.59 & 30.27 & 13.01 & 11.83 & 15.51 & 16.30 & 12.10 & 8.24 & 11.25 & 21.48 & 16.51 & 22.08 & 17.79 \\

& SAR & 31.73 & 28.69 & 38.40 & 13.37 & 35.29 & 15.29 & 14.01 & 20.21 & 21.09 & 14.99 & 10.47 & 14.20 & 24.81 & 22.96 & 28.62 & 22.28 \\

& ROID & 25.48 & 22.97 & 32.32 & 12.42 & 30.22 & 13.24 & 11.81 & 15.50 & 16.94 & 12.76 & 9.07 & 13.49 & 21.43 & 17.74 & 23.13 & 18.57 \\

& DeYO & 27.33 & 24.18 & 32.88 & 12.14 & 32.55 & 13.36 & 12.48 & 16.46 & 17.56 & 13.28 & 8.72 & 11.94 & 22.58 & 18.61 & 24.71 & 19.25 \\

& CMF & 27.31 & 25.02 & 34.56 & 12.06 & 32.42 & 13.64 & 12.30 & 16.95 & 17.87 & 13.42 & 9.06 & 14.07 & 22.41 & 18.82 & 24.84 & 19.65 \\

& Buffer & 32.66 & 29.31 & 38.69 & 12.95 & 35.69 & 14.88 & 13.62 & 18.80 & 19.95 & 14.67 & 9.96 & 14.01 & 24.42 & 21.75 & 28.20 & 21.97 \\

& SPeaR & \textbf{23.68} & \textbf{21.23} & \textbf{30.43} & \textbf{10.94} & \textbf{30.09} & \textbf{12.64} & \textbf{10.74} & \textbf{14.54} & \textbf{14.83} & \textbf{11.22} & \textbf{7.01} & \textbf{9.32} & \textbf{19.88} & \textbf{15.90} & \textbf{21.42} & \textbf{16.93} \\

\bottomrule
\end{tabular}
}
\end{table*}


\begin{table*}[t]
\centering
\caption{Per-corruption error (\%) at severity~5 on CIFAR-100-C, batch sizes $4$, $16$, $256$.}
\label{tab:c100c}
\setlength{\tabcolsep}{5pt}
\resizebox{\textwidth}{!}{%
\renewcommand{\arraystretch}{0.9} 
\begin{tabular}{cl ccc cccc cccc cccc >{\columncolor{summaryblue}}c}
\toprule
& & \multicolumn{3}{c}{Noise} 
& \multicolumn{4}{c}{Blur} 
& \multicolumn{4}{c}{Weather} 
& \multicolumn{4}{c}{Digital} 
& \\
\cmidrule(lr){3-5} \cmidrule(lr){6-9} \cmidrule(lr){10-13} \cmidrule(lr){14-17}
BS & Method 
& GN & SN & IN  
& DB  & GB  & MB  & ZB  
& Snow & Frost & Fog & Bright 
& Cont & Elastic  & Pixel & JPEG
& Mean \\
\midrule
\multirow{6}{*}{%
  \raisebox{-10.ex}{%
    \rotatebox[origin=c]{0}{\small 4}%
  }%
}
& Source & 73.00 & 68.01 & 39.37 & 29.32 & 54.11 & 30.81 & 28.76 & 39.49 & 45.81 & 50.30 & 29.53 & 55.10 & 37.23 & 74.69 & 41.25 & 46.45
\\
\midrule
& TENT & 95.02 & 95.44 & 94.79 & 93.77 & 95.64 & 93.47 & 93.74 & 94.29 & 95.52 & 94.33 & 92.97 & 95.22 & 95.01 & 94.37 & 96.01 & 94.64 \\

& EATA & 90.57 & 89.79 & 89.78 & 90.34 & 91.45 & 88.88 & 90.64 & 90.96 & 91.19 & 89.01 & 89.93 & 91.12 & 91.81 & 89.14 & 87.68 & 90.15 \\

& SAR & 54.96 & 52.56 & 48.82 & 43.26 & 56.49 & 45.00 & 45.27 & 49.02 & 48.39 & 52.46 & 41.07 & 44.98 & 53.40 & 46.04 & 54.08 & 49.05 \\

& ROID & 60.70 & 58.08 & 55.32 & 45.84 & 60.72 & 47.25 & 47.11 & 51.84 & 51.59 & 56.69 & 44.31 & 47.51 & 56.44 & 50.30 & 58.80 & 52.83 \\

& DeYO & 60.47 & 59.07 & 54.32 & 47.19 & 62.41 & 49.39 & 51.65 & 54.46 & 53.56 & 58.68 & 45.22 & 49.98 & 57.53 & 51.62 & 59.64 & 54.35 \\

& CMF & 94.35 & 94.40 & 93.23 & 92.08 & 95.01 & 92.68 & 92.73 & 93.59 & 94.50 & 93.41 & 91.79 & 94.89 & 94.05 & 93.08 & 94.65 & 93.63 \\

& Buffer & 55.28 & 52.91 & 47.75 & 42.23 & 55.43 & 43.12 & 44.76 & 46.98 & 47.88 & 51.59 & 40.67 & 44.43 & 51.99 & 44.52 & 54.24 & 48.25 \\

& SPeaR & \textbf{48.52} & \textbf{46.25} & \textbf{44.21} & \textbf{35.37} & \textbf{50.34} & \textbf{37.24} & \textbf{36.76} & \textbf{41.15} & \textbf{41.09} & \textbf{47.04} & \textbf{33.66} & \textbf{36.83} & \textbf{45.30} & \textbf{38.60} & \textbf{46.59} & \textbf{41.93} \\

\midrule

\multirow{6}{*}{%
  \raisebox{-5.ex}{%
    \rotatebox[origin=c]{0}{\small 16}%
  }%
}
& TENT & 63.02 & 59.84 & 56.17 & 40.10 & 60.65 & 48.67 & 44.80 & 50.42 & 58.76 & 55.96 & 37.83 & 62.29 & 57.53 & 43.82 & 61.86 & 53.45 \\

& EATA & 54.10 & 52.15 & 42.82 & 35.66 & 57.61 & 39.44 & 35.98 & 42.95 & 42.94 & 50.66 & 33.08 & 40.44 & 45.43 & 39.12 & 50.81 & 44.21 \\

& SAR & \textbf{39.62} & \textbf{38.26} & \textbf{33.07} & 28.80 & \textbf{40.49} & 30.79 & 29.69 & \textbf{34.16} & 34.46 & \textbf{36.87} & 27.09 & \textbf{30.53} & \textbf{37.14} & 31.40 & 39.42 & \textbf{34.12} \\

& ROID & 40.82 & 38.52 & 34.71 & 29.29 & 41.45 & 30.98 & 29.79 & 34.39 & 34.94 & 37.22 & 27.57 & 32.13 & 37.57 & 33.17 & 40.49 & 34.87 \\

& DeYO & 41.71 & 39.29 & 36.17 & 29.46 & 41.55 & 31.29 & 29.81 & 35.51 & 35.27 & 39.78 & 28.06 & 31.40 & 37.97 & 33.27 & 40.87 & 35.43 \\

& CMF & 55.79 & 53.16 & 45.95 & 32.30 & 51.19 & 39.29 & 36.80 & 39.51 & 49.10 & 44.06 & 30.52 & 44.56 & 46.07 & 37.37 & 52.64 & 43.89 \\

& Buffer & 41.03 & 38.87 & 35.43 & 29.33 & 41.36 & 30.68 & 29.61 & 33.74 & 34.99 & 36.12 & 27.78 & 35.54 & 37.57 & 31.29 & 40.63 & 34.93 \\

& SPeaR & 40.16 & 38.65 & 36.23 & \textbf{27.71} & 40.85 & \textbf{30.10} & \textbf{28.37} & 34.73 & \textbf{33.91} & 38.26 & \textbf{26.41} & 30.86 & 37.78 & \textbf{30.69} & \textbf{39.04} & 34.25 \\

\midrule

\multirow{6}{*}{%
  \raisebox{-4.ex}{%
    \rotatebox[origin=c]{0}{\small 256}%
  }%
}

& TENT & 37.88 & 35.64 & 35.99 & 26.09 & 37.54 & 27.49 & 25.69 & 31.04 & 32.90 & 34.39 & 24.72 & 28.37 & 33.57 & 29.30 & 37.13 & 31.85 \\

& EATA & 37.63 & 35.90 & 32.12 & 25.88 & 37.91 & 27.81 & 26.05 & 31.45 & 31.01 & 34.22 & 24.78 & 27.25 & 33.86 & 28.84 & 37.10 & 31.45 \\

& SAR & 41.18 & 39.35 & 40.27 & 27.35 & 40.93 & 29.33 & 27.70 & 34.46 & 34.16 & 40.11 & 26.42 & 29.89 & 35.28 & 32.07 & 40.32 & 34.59 \\

& ROID & 37.02 & \textbf{35.41} & 35.02 & 25.65 & 36.93 & 27.09 & 25.90 & \textbf{30.30} & \textbf{30.58} & \textbf{33.60} & 23.99 & 27.26 & \textbf{32.72} & 29.09 & 36.64 & 31.15 \\

& DeYO & 41.89 & 40.11 & 42.15 & 27.50 & 41.55 & 29.56 & 28.04 & 34.97 & 34.46 & 40.95 & 26.58 & 30.03 & 35.58 & 32.74 & 40.76 & 35.12 \\

& CMF & 37.75 & 35.57 & 36.11 & 26.07 & 37.60 & 27.50 & 25.69 & 31.03 & 32.76 & 34.42 & 24.55 & 28.44 & 33.57 & 29.23 & 37.07 & 31.82 \\

& Buffer & 39.46 & 37.53 & 39.28 & 27.02 & 39.76 & 28.90 & 27.02 & 32.68 & 32.47 & 38.83 & 25.42 & 29.54 & 33.97 & 30.83 & 38.90 & 33.44 \\

& SPeaR & \textbf{36.50} & 35.48 & \textbf{30.10} & \textbf{25.15} & \textbf{36.90} & \textbf{26.96} & \textbf{25.37} & 31.19 & 30.83 & 34.36 & \textbf{23.63} & \textbf{27.23} & 33.53 & \textbf{27.99} & \textbf{35.06} & \textbf{30.69} \\\\

\bottomrule
\end{tabular}
}
\end{table*}



\begin{table*}[t]
\centering
\caption{Per-corruption error (\%) at severity~5 on ImageNet-C, batch sizes $4$, $16$, $256$. }
\label{tab:inc}
\setlength{\tabcolsep}{5pt}
\resizebox{\textwidth}{!}{%
\renewcommand{\arraystretch}{0.9} 
\begin{tabular}{cl ccc cccc cccc cccc >{\columncolor{summaryblue}}c}
\toprule
& & \multicolumn{3}{c}{Noise} 
& \multicolumn{4}{c}{Blur} 
& \multicolumn{4}{c}{Weather} 
& \multicolumn{4}{c}{Digital} 
& \\
\cmidrule(lr){3-5} \cmidrule(lr){6-9} \cmidrule(lr){10-13} \cmidrule(lr){14-17}
BS & Method 
& GN & SN & IN  
& DB  & GB  & MB  & ZB  
& Snow & Frost & Fog & Bright 
& Cont & Elastic  & Pixel & JPEG
& Mean \\
\midrule



\multirow{6}{*}{%
  \raisebox{-10.ex}{%
    \rotatebox[origin=c]{0}{\small 4}%
  }%
}

& Source & 97.00 & 96.30 & 97.35 & 82.08 & 90.24 & 85.31 & 77.53 & 83.40 & 76.94 & 75.97 & 40.82 & 94.60 & 83.46 & 79.09 & 67.36 & 81.83

\\
\midrule
& TENT & 89.87 & 88.98 & 88.98 & 91.65 & 91.07 & 82.15 & 75.68 & 75.77 & 77.86 & 65.16 & 51.47 & 89.10 & 70.02 & 66.16 & 73.26 & 78.48 \\

& EATA & 92.03 & 91.70 & 91.99 & 93.40 & 93.58 & 85.68 & 78.74 & 77.83 & 78.97 & 68.86 & 54.46 & 90.71 & 73.39 & 68.58 & 76.00 & 81.06 \\

& SAR & 92.24 & 91.22 & 91.23 & 93.21 & 93.10 & 85.39 & 79.03 & 78.87 & 80.53 & 70.82 & 52.52 & 90.64 & 73.50 & 70.82 & 77.64 & 81.38 \\

& ROID & 89.91 & 88.65 & 88.67 & 91.45 & 91.31 & 81.29 & 75.24 & 74.40 & 76.69 & 64.54 & 51.88 & 87.46 & 70.33 & 64.87 & 71.73 & 77.89 \\

& DeYO & 92.29 & 91.21 & 91.12 & 93.31 & 93.12 & 85.20 & 78.81 & 78.72 & 79.76 & 70.33 & 55.17 & 90.78 & 73.91 & 70.58 & 76.97 & 81.42 \\

& CMF & 90.07 & 90.52 & 89.24 & 93.44 & 94.16 & 87.08 & 74.96 & 73.53 & 79.97 & 63.35 & 51.91 & 97.32 & 68.53 & 64.60 & 70.54 & 79.28 \\

& Buffer & 99.48 & 98.75 & 99.29 & 99.74 & 99.11 & 96.30 & 87.38 & 68.95 & 95.18 & 60.05 & 46.10 & 99.65 & 64.56 & 61.24 & 73.35 & 83.28 \\

& SPeaR & \textbf{78.09} & \textbf{75.93} & \textbf{76.02} & \textbf{85.37} & \textbf{81.16} & \textbf{77.13} & \textbf{67.69} & \textbf{60.68} & \textbf{63.46} & \textbf{47.71} & \textbf{33.12} & \textbf{84.71} & \textbf{53.16} & \textbf{49.63} & \textbf{54.61} & \textbf{65.90} \\

\midrule


\multirow{6}{*}{%
  \raisebox{-5.ex}{%
    \rotatebox[origin=c]{0}{\small 16}%
  }%
}

& TENT & 77.65 & 75.55 & 76.77 & 79.22 & 79.32 & 67.46 & 57.30 & 59.33 & 63.94 & 48.07 & 35.86 & 78.50 & 51.26 & 47.10 & 54.42 & 63.45 \\

& EATA & 76.71 & 72.92 & 75.22 & 82.59 & 80.26 & 64.97 & 57.28 & 57.88 & 62.93 & 47.98 & 37.71 & 70.85 & 51.79 & 47.78 & 53.02 & 62.66 \\

& SAR & 84.32 & 82.08 & 83.61 & 85.29 & 85.18 & 69.96 & 57.54 & 60.36 & 64.39 & 48.43 & 36.06 & 76.55 & 51.82 & 47.47 & 54.97 & 65.87 \\

& ROID & 73.13 & \textbf{70.18} & 72.63 & \textbf{75.32} & \textbf{74.44} & \textbf{63.35} & 55.19 & 56.16 & 61.74 & 46.25 & 35.54 & \textbf{67.49} & 48.48 & 44.73 & 51.66 & \textbf{59.75} \\

& DeYO & 79.87 & 77.43 & 81.09 & 86.85 & 86.01 & 70.16 & 57.90 & 58.41 & 63.56 & 46.25 & 35.37 & 82.91 & 50.51 & 45.84 & 54.00 & 65.08 \\

& CMF & 76.30 & 73.54 & 74.82 & 77.78 & 77.71 & 65.73 & 56.17 & 57.89 & 63.16 & 46.97 & 35.57 & 78.55 & 50.02 & 46.47 & 53.08 & 62.25 \\

& Buffer & 96.86 & 96.26 & 97.57 & 97.86 & 97.41 & 94.06 & \textbf{53.27} & \textbf{51.54} & 83.76 & \textbf{41.46} & 30.49 & 99.07 & \textbf{41.58} & \textbf{43.12} & 55.06 & 71.96 \\

& SPeaR & \textbf{72.90} & 71.84 & \textbf{71.81} & 78.17 & 74.93 & 71.75 & 62.16 & 57.38 & \textbf{59.42} & \textbf{43.68} & \textbf{29.50} & 79.65 & 46.65 & 44.42 & \textbf{49.44} & 60.91 \\

\midrule


\multirow{6}{*}{%
  \raisebox{-5.ex}{%
    \rotatebox[origin=c]{0}{\small 256}%
  }%
}

& TENT & 76.20 & 74.76 & 75.30 & 77.77 & 78.18 & 65.22 & 54.01 & 58.06 & 61.62 & 45.12 & 32.91 & 74.39 & 48.94 & 44.36 & 51.33 & 61.21 \\

& EATA & 70.59 & 68.98 & 69.10 & 72.13 & 72.98 & 60.50 & 53.32 & 55.52 & 60.15 & 45.70 & 35.84 & 63.83 & 48.35 & 44.73 & 50.70 & 58.16 \\

& SAR & 74.91 & 74.02 & 73.65 & 76.68 & 77.28 & 64.89 & 54.22 & 58.11 & 61.28 & 45.31 & 33.02 & 70.80 & 49.46 & 44.76 & 51.68 & 60.67 \\

& ROID & 70.67 & 67.88 & 69.82 & 72.48 & 72.57 & \textbf{59.27} & 50.31 & 52.52 & 57.74 & 42.13 & 31.77 & \textbf{63.49} & 44.14 & 40.55 & 47.04 & 56.16 \\

& DeYO & 76.39 & 74.82 & 76.25 & 79.71 & 79.16 & 64.77 & 53.62 & 56.60 & 60.50 & 43.63 & 32.19 & 70.53 & 47.84 & 42.95 & 50.39 & 60.62 \\

& CMF & 76.88 & 75.42 & 75.98 & 78.31 & 78.71 & 65.81 & 54.39 & 58.69 & 62.00 & 45.57 & 32.99 & 75.08 & 49.50 & 44.78 & 51.90 & 61.73 \\

& Buffer & 83.79 & 76.23 & 77.57 & 87.83 & 89.93 & 53.90 & \textbf{46.95} & \textbf{46.53} & 62.19 & \textbf{39.17} & 30.83 & 93.45 & 41.12 & \textbf{38.34} & \textbf{44.84} & 60.84 \\

& SPeaR & \textbf{68.32} & \textbf{65.88} & \textbf{65.00} & \textbf{71.30} & \textbf{68.53} & 62.53 & 54.06 & 50.61 & \textbf{53.97} & 39.31 & \textbf{27.64} & 68.99 & \textbf{39.69} & 39.44 & 46.73 & \textbf{54.80} \\

\bottomrule
\end{tabular}
}
\end{table*}

\section{Group-Normalized and Normalization-Free Backbones}
\label{sec:full_result_2}


Since SPeaR steers intermediate representations through a primitive external to the frozen backbone, its architecture-agnostic design allows the same adaptation mechanism to be applied across architectures with different normalization schemes. We therefore evaluate it on two additional ImageNet backbones: a group-normalized ResNet-50-GN and the normalization-free NFNet-F0. The former enables comparison with methods that support group normalization, while the latter provides a setting without normalization layers, where Buffer is the only applicable baseline.

Tables~\ref{tab:altbb_gn} and \ref{tab:altbb_nonorm} report corruption error at severity~5 for both backbones. On the group-normalized model, SPeaR uses the same steering primitive and adaptation principle and attains the lowest error on every corruption, ahead of all applicable baselines. On the normalization-free backbone, SPeaR delivers consistent improvements over both Buffer and the unadapted source model across every corruption. Together, these results show that SPeaR's improvements are not tied to updating the model's internal parameters or to any particular normalization scheme, supporting its architecture-agnostic design.


\begin{table*}[t]
\centering
\caption{Corruption error (\%) at severity~5 on ImageNet-C for a group-normalized ResNet-50, batch size $256$.}
\label{tab:altbb_gn}
\setlength{\tabcolsep}{5pt}
\resizebox{\textwidth}{!}{%
\renewcommand{\arraystretch}{0.9} 
\begin{tabular}{l ccc cccc cccc cccc >{\columncolor{summaryblue}}c}
\toprule
& \multicolumn{3}{c}{Noise} 
& \multicolumn{4}{c}{Blur} 
& \multicolumn{4}{c}{Weather} 
& \multicolumn{4}{c}{Digital} 
& \\
\cmidrule(lr){2-4} \cmidrule(lr){5-8} \cmidrule(lr){9-12} \cmidrule(lr){13-16}
Method 
& GN & SN & IN  
& DB  & GB  & MB  & ZB  
& Snow & Frost & Fog & Bright 
& Cont & Elastic  & Pixel & JPEG
& Mean \\
\midrule
Source & 72.29 & 71.87 & 71.23 & 82.01 & 90.42 & 76.50 & 73.38 & 57.77 & 53.94 & 64.36 & 29.25 & 62.02 & 85.75 & 84.85 & 46.47 & 68.14 \\ 
SAR & 68.94 & 66.80 & 66.76 & 82.17 & 88.48 & 73.58 & 72.18 & 55.69 & 57.92 & 53.81 & 28.94 & 53.98 & 87.21 & 61.49 & 45.47 & 64.23 \\
ROID & 62.87 & 60.69 & 61.04 & 81.17 & 82.44 & 69.57 & 66.76 & 49.91 & 50.46 & 43.90 & 27.35 & 48.33 & 68.43 & 54.58 & 43.73 & 58.08 \\
DeYO & 72.76 & 70.97 & 71.41 & 82.00 & 89.83 & 75.73 & 73.58 & 57.46 & 54.48 & 66.15 & 29.26 & 58.43 & 84.95 & 75.01 & 46.15 & 67.21 \\
Buffer & 68.60 & 68.97 & 66.98 & 84.13 & 83.55 & 71.51 & 74.03 & 55.62 & 71.41 & 51.10 & 29.57 & 52.72 & 92.98 & 57.19 & 46.06 & 64.96 \\
SPeaR & \textbf{62.15} & \textbf{59.46} & \textbf{60.14} & \textbf{70.98} & \textbf{72.76} & \textbf{57.56} & \textbf{55.60} & \textbf{43.20} & \textbf{49.07} & \textbf{37.20} & \textbf{26.17} & \textbf{42.74} & \textbf{51.85} & \textbf{42.49} & \textbf{41.02} & \textbf{51.49} \\
\bottomrule
\end{tabular}
}
\end{table*}


\begin{table*}[t]
\centering
\caption{Corruption error (\%) at severity~5 on ImageNet-C for normalization-free NFNet, batch size $256$. Methods that adapt normalization parameters do not directly apply.}
\label{tab:altbb_nonorm}
\setlength{\tabcolsep}{5pt}
\resizebox{\textwidth}{!}{%
\renewcommand{\arraystretch}{0.9} 
\begin{tabular}{l ccc cccc cccc cccc >{\columncolor{summaryblue}}c}
\toprule
& \multicolumn{3}{c}{Noise} 
& \multicolumn{4}{c}{Blur} 
& \multicolumn{4}{c}{Weather} 
& \multicolumn{4}{c}{Digital} 
& \\
\cmidrule(lr){2-4} \cmidrule(lr){5-8} \cmidrule(lr){9-12} \cmidrule(lr){13-16}
Method 
& GN & SN & IN  
& DB  & GB  & MB  & ZB  
& Snow & Frost & Fog & Bright 
& Cont & Elastic  & Pixel & JPEG
& Mean \\
\midrule
Source & 69.53 & 66.22 & 71.90 & 75.48 & 86.91 & 66.84 & 66.50 & 59.58 & 47.03 & 69.40 & 27.49 & 84.79 & 77.76 & 64.92 & 39.09 & 64.90 \\
Buffer & 66.22 & 60.84 & 65.00 & 76.54 & 84.02 & 65.76 & 65.61 & 56.77 & 49.30 & 93.01 & 27.80 & 84.46 & 78.44 & 61.35 & 38.54 & 64.91 \\
SPeaR & \textbf{49.69} & \textbf{47.93} & \textbf{48.69} & \textbf{61.16} & \textbf{60.36} & \textbf{51.28} & \textbf{50.85} & \textbf{42.75} & \textbf{42.81} & \textbf{33.98} & \textbf{25.42} & \textbf{40.22} & \textbf{37.00} & \textbf{32.16} & \textbf{35.07} & \textbf{43.96} \\
\bottomrule
\end{tabular}
}
\end{table*}


\section{Results on Vision Transformer}
\label{sec:full_result_3}

Since the steering primitive is defined at stage boundaries, the same implementation used for CNNs is applied directly to ViT-B/16. Table~\ref{tab:vit} extends Table~2 of the main paper into the full per-corruption breakdown at severity~5, evaluated at batch sizes $4$, $16$, and $256$ against SAR, ROID, and DeYO. At batch sizes 16 and 256, SPeaR achieves the lowest mean error among all methods while remaining competitive across individual corruption types. At batch size 4, it remains competitive with the strongest baseline. Its per-corruption behavior is also markedly stable: on corruptions where the baselines become unstable, SPeaR continues to adapt reliably, which is a further source of its strong mean. These results further demonstrate that the same steering primitive remains effective across both convolutional and Transformer backbones.

\begin{table*}[t]
\centering
\caption{Per-corruption error (\%) at severity~5 on ImageNet-C with ViT-B/16, batch sizes $4$, $16$, $256$.}
\label{tab:vit}
\setlength{\tabcolsep}{5pt}
\resizebox{\textwidth}{!}{%
\renewcommand{\arraystretch}{0.9} 
\begin{tabular}{cl ccc cccc cccc cccc >{\columncolor{summaryblue}}c}
\toprule
& & \multicolumn{3}{c}{Noise} 
& \multicolumn{4}{c}{Blur} 
& \multicolumn{4}{c}{Weather} 
& \multicolumn{4}{c}{Digital} 
& \\
\cmidrule(lr){3-5} \cmidrule(lr){6-9} \cmidrule(lr){10-13} \cmidrule(lr){14-17}
BS & Method 
& GN & SN & IN  
& DB  & GB  & MB  & ZB  
& Snow & Frost & Fog & Bright 
& Cont & Elastic  & Pixel & JPEG
& Mean \\
\midrule

\multirow{6}{*}{%
  \raisebox{-2.ex}{%
    \rotatebox[origin=c]{0}{\small 4}%
  }%
}

& Source & 50.32 & 49.78 & 49.97 & 57.29 & 65.63 & 49.44 & 55.25 & 43.45 & 47.69 & 43.36 & 24.11 & 68.05 & 53.11 & 34.47 & 33.61 & 48.37

\\
\midrule

& SAR & \textbf{38.59} & \textbf{36.94} & \textbf{37.53} & \textbf{40.87} & \textbf{36.83} & \textbf{30.17} & \textbf{31.63} & \textbf{26.10} & \textbf{29.40} & \textbf{25.13} & \textbf{18.11} & 99.91 & \textbf{22.43} & \textbf{21.71} & \textbf{25.13} & \textbf{34.70} \\

& ROID & 41.63 & 40.07 & 40.39 & 43.87 & 42.29 & 36.91 & 42.06 & 30.60 & 33.65 & 45.56 & 19.09 & 99.85 & 27.32 & 25.05 & 27.62 & 39.73 \\

& DeYO & 95.93 & 92.14 & 97.91 & 99.76 & 93.81 & 99.84 & 99.39 & 94.87 & 95.44 & 99.53 & 24.28 & 99.99 & 96.79 & 55.98 & 46.01 & 86.11 \\

& SPeaR & 41.84 & 40.19 & 40.89 & 43.91 & 41.86 & 38.02 & 42.97 & 31.91 & 35.39 & 32.56 & 24.54 & \textbf{40.31} & 30.46 & 27.15 & 30.45 & 36.16 \\

\midrule

\multirow{6}{*}{%
  \raisebox{4.ex}{%
    \rotatebox[origin=c]{0}{\small 16}%
  }%
}

& SAR & \textbf{36.00} & \textbf{34.87} & \textbf{35.17} & \textbf{36.04} & 36.05 & 30.30 & \textbf{33.02} & 26.46 & \textbf{28.08} & \textbf{24.51} & \textbf{17.04} & 99.17 & 24.02 & 21.57 & \textbf{23.97} & 33.75 \\

& ROID & 37.82 & 36.79 & 37.21 & 38.24 & 40.13 & 33.97 & 37.08 & 28.66 & 30.52 & 25.61 & 17.47 & 39.39 & 27.08 & 23.75 & 26.06 & 31.99 \\

& DeYO & 42.93 & 35.03 & 37.09 & 50.35 & 36.41 & \textbf{31.25} & 96.74 & \textbf{25.84} & 31.96 & 27.68 & 17.61 & 99.95 & \textbf{22.25} & \textbf{20.91} & 24.89 & 40.06 \\

& SPeaR & 38.52 & 35.70 & 35.99 & 37.89 & \textbf{35.46} & 31.84 & 34.23 & 26.99 & 30.51 & 26.76 & 22.74 & \textbf{33.03} & 26.14 & 24.73 & 26.47 & \textbf{31.13} \\

\midrule

\multirow{6}{*}{%
  \raisebox{3.ex}{%
    \rotatebox[origin=c]{0}{\small 256}%
  }%
}

& SAR & 41.84 & 40.66 & 40.79 & 42.50 & 45.61 & 39.06 & 42.34 & 34.39 & 35.90 & 32.49 & 21.27 & 38.42 & 37.44 & 27.60 & 29.60 & 36.66 \\

& ROID & 41.55 & 40.39 & 40.62 & 42.13 & 45.78 & 38.56 & 41.96 & 33.46 & 34.95 & 30.42 & 20.86 & 37.53 & 36.25 & 27.17 & 29.40 & 36.07 \\

& DeYO & 39.38 & 37.71 & 37.98 & \textbf{40.12} & 39.85 & 34.87 & 36.60 & 29.93 & 32.03 & 27.50 & 20.24 & 99.63 & 29.37 & 24.36 & 26.88 & 37.10 \\

& SPeaR & \textbf{39.22} & \textbf{37.33} & \textbf{37.82} & 40.55 & \textbf{39.07} & \textbf{33.79} & \textbf{34.87} & \textbf{28.86} & \textbf{30.96} & \textbf{26.38} & \textbf{20.05} & \textbf{33.08} & \textbf{28.02} & \textbf{24.13} & \textbf{26.70} & \textbf{32.06} \\

\bottomrule
\end{tabular}
}
\end{table*}

\section{Parameter Efficiency}
\label{p_v_p}

\begin{figure}[t]
    \centering
    \includegraphics[width=\linewidth]{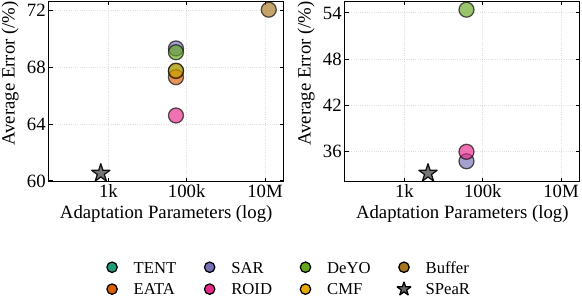}
\caption{State-of-the-art comparison of performance (percentage error, lower is better) versus the number of adapted parameters on ImageNet-C on ResNet-50 and ViT-B/16, averaged across batch sizes.}
\label{fig:param-perf}
\end{figure}

Figure~\ref{fig:param-perf} extends the parameter-versus-performance analysis of Figure~2 to the two ImageNet-C backbones, plotting average error against the number of adaptation parameters on a logarithmic scale. On both ResNet-50 and ViT-B/16, SPeaR achieves the lowest average error while adapting the fewest parameters. Adapting more parameters does not necessarily yield better performance. This confirms that the efficiency advantage seen in the main paper persists at ImageNet scale and across both convolutional and Transformer backbones, and that competitive test-time adaptation does not require a large adaptation budget.

\section{Hard Gating Analysis}

\begin{figure}[t]
    \centering
    \includegraphics[width=.7\linewidth]{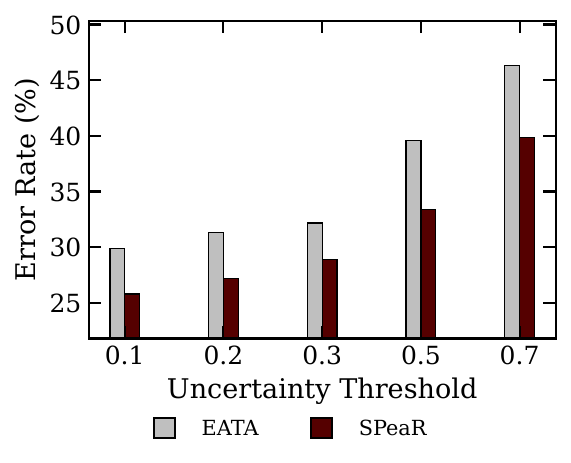}
    \caption{SPeaR versus EATA on CIFAR-10-C as confident samples grow scarce. Only images with source-model entropy above a threshold are used for adaptation, and the threshold is raised progressively.}
    \label{fig:eata_comp}
\end{figure}

\begin{figure*}[t]
\centering
\includegraphics[width=.80\textwidth]{"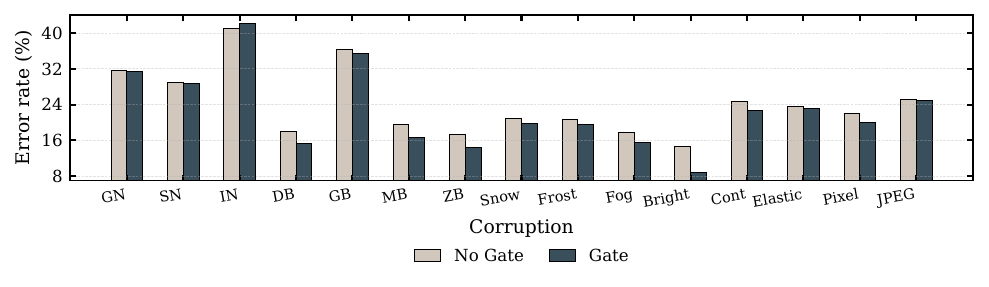"}
\caption{Hard Gate ablation study on CIFAR-10-C across all corruptions (severity 5). The benefit of the hard gate concentrates on lower-entropy corruptions (brightness, the blur types, fog), where predictions are often already confident, and is negligible on high-entropy corruptions.}
\label{fig:hard_gate_abb}
\end{figure*}

Entropy minimization is the standard objective in test-time adaptation, and EATA \cite{niu2022efficient} refines it with a sample-selection rule that upweights confident, low-entropy predictions as the more reliable signal. Applied to a full-network update, this is effective, and EATA remains a strong baseline throughout our experiments.

Our hard gate instead withholds the entropy objective from batches whose mean entropy falls below a predefined threshold. This differs from EATA because the two mechanisms operate on different adaptation components. EATA adapts the BN layers throughout the network, whereas our hard gate governs a low-dimensional steering primitive. We find that this compact space is sensitive to repeated pressure from already-confident predictions: once a batch is confident, further entropy minimization yields little correction but keeps moving the primitive. Skipping such batches leaves the primitive untouched.

\noindent\textbf{Hard-gate ablation.} Figure~F.4 reports corruption error with and without the gate on CIFAR-10-C. Averaged over all corruptions the gain is small, because on high-entropy corruptions few predictions are confident enough to be gated and the two variants behave alike. The gain concentrates where expected: on lower-entropy corruptions such as brightness, the blur types, and fog, the gate lowers error by a clear margin.

\noindent\textbf{Behavior when confident samples are scarce.} We progressively remove the most confident images from CIFAR-10-C, forcing adaptation to rely on increasingly uncertain samples. As expected, the corruption error increases for both SPeaR and EATA as fewer reliable samples remain. However, SPeaR consistently achieves lower error, with the gap widening as the available samples become more uncertain. This suggests that SPeaR is less dependent on confident samples for effective adaptation. EATA remains a strong baseline when confident samples are naturally present in the test stream.



\section{Implementation Details}
\label{sec:hyperparameters}

SPeaR is tuned once per dataset, with the resulting learning rate, adaptation steps, and loss weights reused across all batch sizes and corruption types. The complete implementation and hyperparameter configurations will be made publicly available upon publication.


\clearpage

\end{document}